%% file: ms.tex
\documentclass[11pt,a4paper]{article}
\usepackage[hyperref]{emnlp2020}
\usepackage{times}
\usepackage{latexsym}

\usepackage{microtype}

\usepackage[toc,page]{appendix}
\usepackage{array}
\usepackage{graphicx}
\usepackage{subfig}
\usepackage{amsmath}
\usepackage{comment}
\usepackage[para]{threeparttable}
\usepackage{booktabs}
\usepackage{makecell}
\usepackage{tabularx}
\usepackage{multirow}

\graphicspath{{figs/}}
\DeclareGraphicsExtensions{.pdf,.png,.jpg}

\DeclareMathOperator*{\argmax}{arg\,max}

\aclfinalcopy 

\newcolumntype{+}{>{\global\let\currentrowstyle\relax}}
\newcolumntype{=}{>{\currentrowstyle}}
\newcommand{\rowstyle}[1]{\gdef\currentrowstyle{#1}#1\ignorespaces}

\newcommand{\pkm}{PKM}
\newcommand{\resm}{ResM}

\newcommand{\bb}{BERT\textsubscript{BASE}}
\newcommand{\bl}{BERT\textsubscript{LARGE}}
\newcommand{\db}{DistilBERT}

\newcommand{\memu}{$MU$}
\newcommand{\tmemu}{$\widetilde{MU}$}
\newcommand{\klu}{$KL_u$}
\newcommand{\klw}{$KL_w$}

\title{Large Product Key Memory for Pretrained Language Models}

\author{
Gyuwan Kim\thanks{\, Equal contribution.} \\
Clova AI, NAVER Corp. \\
\tt{gyuwan.kim@navercorp.com}
\And
Tae-Hwan Jung\footnotemark[1]\,\,\thanks{\, TJ was an intern at Clova AI while doing this work.} \\
Kyung Hee University \\
\tt{nlkey2022@gmail.com}
}

\date{}

\begin{document}
\maketitle

\begin{abstract}
\input{00_abstract}
\end{abstract}

\input{01_introduction}
\input{02_background}
\input{03_catastrophic_drift}
\input{04_pretraining}
\input{05_experiments}
\input{06_pretraining_results}
\input{07_finetuning_results}
\input{08_conclusion_and_future_work}
\input{09_acknowledgements}

\bibliography{emnlp2020}
\bibliographystyle{acl_natbib}

\end{document}

%% file: 00_abstract.tex
Product key memory (PKM) proposed by \citet{lample2019large} enables to improve prediction accuracy by increasing model capacity efficiently with insignificant computational overhead. However, their empirical application is only limited to causal language modeling. Motivated by the recent success of pretrained language models (PLMs), we investigate how to incorporate large PKM into PLMs that can be finetuned for a wide variety of downstream NLP tasks. We define a new memory usage metric, and careful observation using this metric reveals that most memory slots remain outdated during the training of PKM-augmented models. To train better PLMs by tackling this issue, we propose simple but effective solutions: (1) initialization from the model weights pretrained without memory and (2) augmenting PKM by addition rather than replacing a feed-forward network. We verify that both of them are crucial for the pretraining of PKM-augmented PLMs, enhancing memory utilization and downstream performance. Code and pretrained weights are available at \href{https://github.com/clovaai/pkm-transformers}{https://github.com/clovaai/pkm-transformers}.

%% file: 01_introduction.tex
\section{Introduction}

Larger model capacity has brought improvement in accuracy by enabling better modeling of data. 
However, increasing model capacity causes a significant increase in computational cost at both training and inference time despite better accuracy.
To address this issue, \citet{lample2019large} propose product key memory (PKM) that enables very efficient and exact nearest neighbor search in a large number of learnable memory slots.
They substitute a feed-forward network (FFN) in a transformer block \cite{vaswani2017attention} with a PKM layer.
Augmenting large PKM layers to networks allows increasing model capacity, with only a slight increase in inference time.
\citet{lample2019large} prove the efficiency of PKM on causal language models (CLMs) in terms of the superior trade-off between perplexity and inference speed.
For instance, they achieve a PKM-augmented CLM with only 12 layers that is more accurate and twice faster than a baseline with 24 layers.

\input{tabs/speed}

However, usage of PKM with a pretrained language model (PLM) such as BERT \cite{devlin2018bert} that is helpful for downstream tasks \cite{wang2018glue} has not been examined in the literature.
In our experiments, plain PKM improves masked language modeling (MLM) perplexity but not downstream performance.

We measure various memory utilization metrics to analyze how many memory slots contribute to the model prediction.
Careful examination about memory utilization during and after the training demonstrates that only a few memory slots are being used importantly (\S~\ref{sec:castastrophic_drift}).
We attribute this phenomenon, called a catastrophic drift, to the sparsely updated memory parameters.
The lower memory utilization implies that model capacity from memory is not fully exploited.
It promotes us to develop methods that can overcome this issue.

We found that initialization from weights pretrained without memory is essential for pretraining PKM-augmented PLMs.
Moreover, rather than replacing an FFN to a PKM as \citet{lample2019large} do, we show that adding PKM to a transformer layer \cite{vaswani2017attention} with a residual connection \cite{he2016deep} without removing FFN is advantageous.
Both the initialization (\S~\ref{sec:init}) and our proposed residual memory (\resm, \S~\ref{sec:residual}) prevent a sudden change of transformer parameters, thus allow to train memory parameters better by less suffering from the catastrophic drift.
Consequently, we obtain PKM-augmented-\bb{} having comparable accuracy and faster than \bl. 

As demonstrated in Table \ref{tab:speed}, 
a model with a large memory is much faster than a model having twice many transformer layers, although it has far more weights.
\resm{} does not slow down inference speed much.
Accuracy comparison between them will appear in the later sections.

The main contributions of this work are summarized as follows.
First, we explore how to incorporate PKM to PLMs to be finetuned for downstream tasks and find that simple application does not work well.
Secondly, we attribute this to a catastrophic drift during the training by careful monitoring of memory utilization.
Lastly, we propose simple yet effective solutions to tackle the observed catastrophic drift problem: (1) weight initialization without PKM and (2) the residual memory layer.
We empirically verify that both of them are crucial to achieve improved accuracy.
In our knowledge, this is the first work that successfully applies PKM to PLMs.

%% file: tabs/speed.tex
\begin{table}[t]
    \centering
    \footnotesize
    
    \begin{tabular*}{0.48\textwidth}{lrrr}
    \toprule
    \multicolumn{1}{c}{Model} & 
    \multicolumn{1}{c}{\# Layers} & 
    \multicolumn{1}{c}{\# Params} &
    \multicolumn{1}{c}{\makecell{Inference \\ Speed \\ \scriptsize (batch/sec)}}  \\
    \midrule
    \bb & 12    & 110M & 79.8 \\
    \bb~+PKM    & 12 & 506M & 61.4  \\
    \bb~+ResM   & 12 & \textbf{515M} & \textbf{59.3}  \\
    \midrule[0.2pt]
    \bl         & 24 & \textbf{340M} & \textbf{43.1} \\
    \bl~+PKM    & 24 & 860M & 37.2 \\
    \bl~+ResM   & 24 & 876M & 36.1 \\
    \bottomrule
    \end{tabular*}
    \caption{
    \label{tab:speed}
    Comparison of inference speed between different model sizes and the memory layers. 
    We run each model for the classification task with batch size 1, and measure inference speed on a single V100 GPU.
    We follow the model size settings of BERT \cite{devlin2018bert}.
    We use two memory layers with the recommended setting of PKM hyper-parameters following \citet{lample2019large} as described in \textsection \ref{sec:expr-setting}.
    As marked bold, \bb~with our proposed residual memory (ResM) is much faster than \bl, while having more parameters.
    }
\end{table}

%% file: 02_background.tex
\section{Background}

\subsection{Transformers and Product Key Memory}

A transformer encoder maps a sequence of input tokens into a sequence of continuous representations based on a self-attention mechanism~\citep{vaswani2017attention}.
Transformer architecture is a stack of sub-layers, and each sub-layer consists of a multi-head attention layer and a feed-forward layer.
Due to the remarkable prediction accuracy, a transformer becomes standard architecture in natural language processing.

On the other hand, memory architecture can also be used to design a function that maps a continuous representation to another representation as a layer in neural networks.
When a query vector is given in a standard memory-augmented neural network, the memory layer finds $k$-NN keys and returns a weighted sum of corresponding value vectors.
These weights are normalized scores of the dot product between the query vector and the key vectors.

\citet{lample2019large} propose product key memory (PKM) that can significantly increase model capacity based on fast and exact nearest neighbor search.
They plug a PKM layer in a transformer architecture, especially by switching an existing feed-forward layer to it, while keeping similar computational efficiency.

We explain the mechanism of PKM here to be self-contained.
A product key is a pair of sub-keys, meaning that there are $|K| = C^2$ different memory slots when the codebook size of each sub-key is $C$.
A given query vector is partitioned to the dimension of half-size.
The score with a product key is the sum of the dot product between the sub-query vector and the sub-key vector.
We can increase the size of key space effectively with sufficient $C$.
Exact nearest neighbor search in the product key set can be done efficiently by first finding $k$-NN in each sub-key space and then finding $k$-NN again from $k^2$ combinations of sub-key pairs.
In addition, a multi-head memory attention mechanism like self-attention in transformers is used to increase the representation power of the memory layer.

\subsection{Pretrained Language Models}

Transfer learning from pretrained language models (PLMs) has brought a paradigm shift in NLP with a remarkable improvement in a wide range of downstream tasks. 
Based on a transformer architecture \cite{vaswani2017attention}, BERT \cite{devlin2018bert} is trained with two pretraining tasks, (1) masked language modeling (MLM) and (2) next sentence prediction (NSP), which achieves significant improvement in performance on fine-tuning tasks. 
RoBERTa \cite{liu2019roberta} removes the NSP and increases the batch size and training corpus to train a more robust language model. 
It indicates that larger batch size and training data benefit the performance of PLM. 
In these trends, recently, language models with much larger parameters \cite{raffel2019exploring, shoeybi2019megatron, brown2020language} are trained with a huge amount of text corpus. 
Despite their remarkable performance, the computational cost in training and inference is prohibitive.
Improving trade-off between accuracy and efficiency is one of the crucial research directions.

\subsection{Memory-Augmented Language Models}

Memory augmented neural networks \cite{weston2014memory, sukhbaatar2015end} have the ability to solve complex algorithmic tasks and decouple the memory capacity from the number of model parameters.
\citet{chandar2016hierarchical} propose a hierarchical memory network to access from large external memory efficiently.
\citet{rae2016scaling} enable training a large memory in neural networks efficiently via a sparse read and write mechanism.
However, it requires regular re-training to avoid a catastrophic drift.
REALM \cite{guu2020realm} also suffers from a similar issue, so refresh the index asynchronously every several hundred training steps.

In addition to \citet{lample2019large}, augmenting memory architecture to a language model is a promising research direction.
For example, EaE \cite{fevry2020entities} and FaE \cite{verga2020facts} jointly train a memory that is interleaved in a transformer and dedicated to entities (or facts) with sparse updates, and access to only a small portion of the memory in inference time.
On the other hand, each memory slot in \citet{lample2019large} and ours does not have explicit meaning.

\citet{sukhbaatar2019augmenting} augments the self-attention layers with persistent memory vectors and removes the feed-forward layers.
\citet{khandelwal2019generalization} augments a pretrained language model with the nearest neighbor language model that retrieves $k$-nearest neighbors from the datastore consisting of the key-value pairs of a context vector and the target word built from training data.
\citet{khandelwal2019generalization} also only considers causal language modeling, and applying the same approach to masked language modeling widely used for PLMs is non-trivial.

%% file: 03_catastrophic_drift.tex
\section{Memory Utilization Analysis}

As shown in our experiment (Table~\ref{tab:pretraining}), large PKM provides a significant gain in masked language modeling in terms of perplexity. 
However, surprisingly, downstream task performance finetuned from PKM-augmented PLMs is similar to or sometimes worse than that without PKM in our experiments.
Nevertheless, it is challenging to investigate what is going on under the hood.
We presume that this frustrating outcome come from the catastrophic drift which will be explained later (\S~\ref{sec:castastrophic_drift}) and it fosters us to scrutinize memory utilization (\S~\ref{sec:memory_utilization}) thoroughly.

\subsection{Catastrophic Drift}
\label{sec:castastrophic_drift}
PKM is jointly trained with other transformer parameters.
In every training step, only a small portion (chosen as $k$-NN) of memory parameters are sparsely updated.
Even if a memory slot is selected as top-$k$, the frequency is low or it is only selected as low-rank in top-$k$, the update of memory parameters relevant to this slot might be marginal.

If memory parameters (especially value vectors) are not updated (or rarely updated) for a while, they became stale.
Stale parameters are unlikely to be matched with newly updated model parameters so that they will get remain unused.
We call this situation a \textit{catastrophic drift}.
Moreover, catastrophic drift will be more severe in finetuning because it relies on a small number of data and training steps.

We hypothesize this catastrophic drift occurs during the training of a PKM-augmented LM, and it is one plausible cause of poor performance.
This problem is overlooked by \citet{lample2019large} because it is concealed by increasing the number of memory slots $|K|$, heads $H$, or $k$-NN.
With a sufficient size of memory hyper-parameters, memory usage (see \S~\ref{sec:memory_utilization} for the definition) becomes close to 100\%.
For example, in \citet{lample2019large} and our experiments, memory usage is almost 100\% when using 4 memory heads, selecting 32 keys per head, and using 512\textsuperscript{2} memory slots.
Considering only top-k memory usage, memory parameters are seemingly regarded as used effectively to their full extent.

\subsection{Memory Utilization Metrics}
\label{sec:memory_utilization}
Following \citet{lample2019large}, we measure the memory utilization of trained PKM-augmented models in terms of (1) memory usage and (2) KL divergence with the uniform distribution using held-out data.
Besides standard memory usage, we propose to measure top-1 memory usage that only counts memory slots as used when selected as top-1 rather than top-k and use it to monitor the degree of catastrophic drift.

For every memory slot, we count the number of selection as k-NN (or top-1) and sum the weights throughout all memory accesses: $u'_i=\sum_x \delta(w(x)_i>0)$, $t'_i=\sum_x \delta(\argmax _j w(x)_j=i)$, and $w'_i=\sum_x w(x)_i$, where $w(x)_i$ is the weight of the key $i$ accessed in the memory when an input $x$ is given to the language model with the memory.
Memory usage ($MU$) is the fraction of values that are accessed at least once.
Top-1 memory usage ($\widetilde{MU}$) is the fraction of values that are accessed as top-1 at least once.
KL divergence with the uniform distribution is calculated for normalized average counts (\klu) and normalized average weights (\klw).
Formally, we can calculate those values by
\begin{equation*}
\begin{aligned}
    MU   &= \frac{1}{|K|} \sum_i \delta(u_i > 0), \\
    \widetilde{MU} &= \frac{1}{|K|} \sum_i \delta(t_i > 0), \\
    KL_u &= log(|K|) + \sum_i u_i log(u_i), \\
    KL_w &= log(|K|) + \sum_i w_i log(w_i) \\
\end{aligned}
\end{equation*}
where $|K|$ is the number of memory slots, and $u$, $t$, and $w$ are the normalized value of $u'$, $t'$, and $w'$, respectively, as sum to 1.

%% file: 04_pretraining.tex
\input{figs/resm}

\section{Pretraining PKM-augmented PLMs}

\citet{lample2019large} propose PKM and show its advantage in causal language modeling.
We investigate how to extend the usage of large PKM to PLMs such as BERT \cite{devlin2018bert} and RoBERTa \cite{liu2019roberta} that can be used as a good initialization point for downstream tasks, resulting in a great performance.

By monitoring top-1 memory usage, we observe that catastrophic drift really occurs.
Low memory utilization PKM-augmented PLMs means that the model does not fully exploit its increased capacity of the memory and thus is likely not to get accuracy gain much.
To resolve the catastrophic drift, we introduce additional modifications for better pretraining: initialization from pretrained weights (\S~\ref{sec:init}) and residual memory layer (\S~\ref{sec:residual}).

\subsection{Initialization from Pretrained Weights} 
\label{sec:init}

Learning transformer parameters and memory parameters together from scratch is difficult due to the discrepancy between them as described in \S~\ref{sec:castastrophic_drift}.
To remedy this issue, we first pretrain a language model without memory layers, and then pretrain again a model with memory layers initialized from the already pretrained language model.
Transformer parameters will be gradually changed since they are initialized from a well-trained language model.
We expect that staleness would be mitigated as a result.
Despite requiring two stages of training, a trained language model with initialization performs much better and has higher memory usage than that with the same amount of training steps from the scratch, as shown in Table \ref{tab:pretraining}.

\subsection{Residual Memory Layer} 
\label{sec:residual}

\citet{he2016deep} propose ResNet to train very deep convolution networks.
A residual connection enables easier optimization and gains accuracy from increased depth.
We borrow this idea by introducing a residual connection in augmenting a PKM to alleviate the catastrophic drift.

When we replace an FFN layer of pretrained networks with the PKM layer, it struggles to fit data in an early stage because the function of this layer suddenly changed to random function from a well-trained one (see a green line of Figure \ref{fig:training_curve}). 
We hope to prevent this undesirable circumstance while keeping strong representation power of product key memory.
To this end, we propose residual memory (ResM) layer, adding the memory layer to a transformer block in the form of residual connection \cite{he2016deep} instead of replacing the FFN layer.
Due to the residual connection, the function of the layer does not deviate severely from that of the original pretrained weights, and it helps to start at a stable point.

Figure \ref{fig:residual_memory_layer} displays how the residual memory layer is different from the previous models.
To be more precise, we can formulate these layers to
\begin{equation*}
x' = LN(x + \alpha FFN(x) + \beta PKM(x)),
\end{equation*}
where $LN$ indicates layer normalization \cite{ba2016layer}.
$(\alpha, \beta) = (1, 0), (0, 1), (1, 1)$ corresponds to FFN layer, PKM layer, and ResM layer, respectively.

%% file: figs/resm.tex
\begin{figure*}[t!]
    \centering 
    \subfloat[FFN layer]{\includegraphics[width=0.23\textwidth]{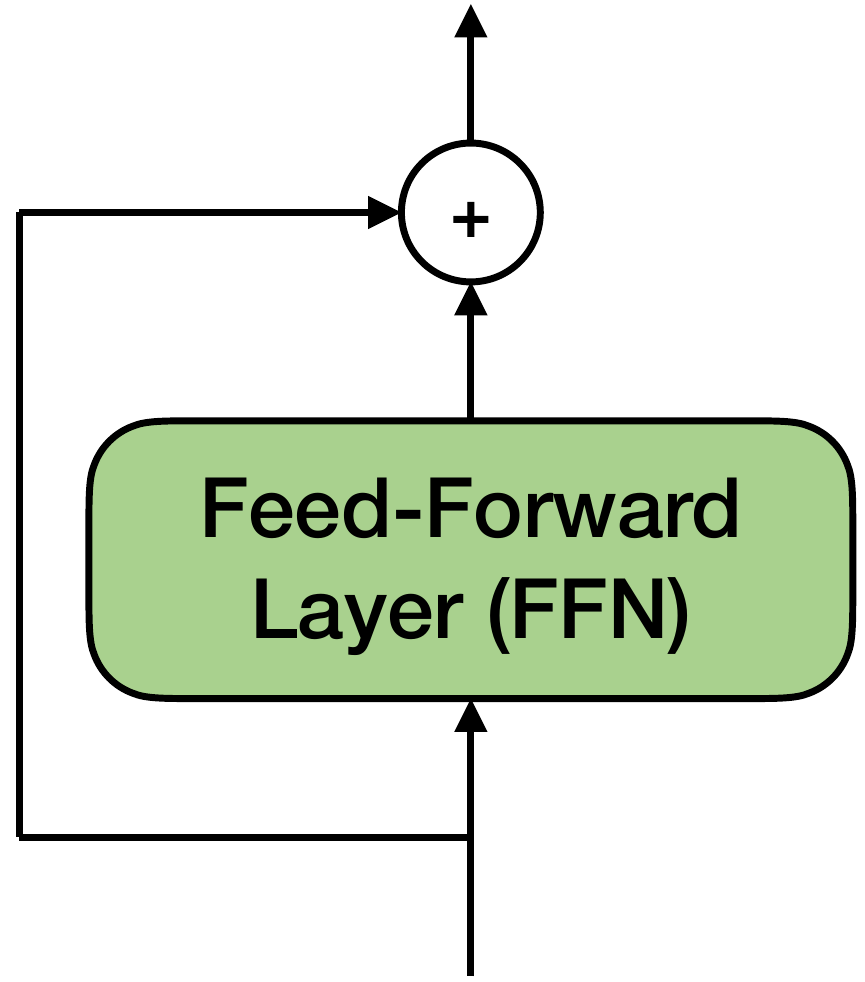}}
    \:\:\:\:
    \subfloat[\pkm~layer]{\includegraphics[width=0.23\textwidth]{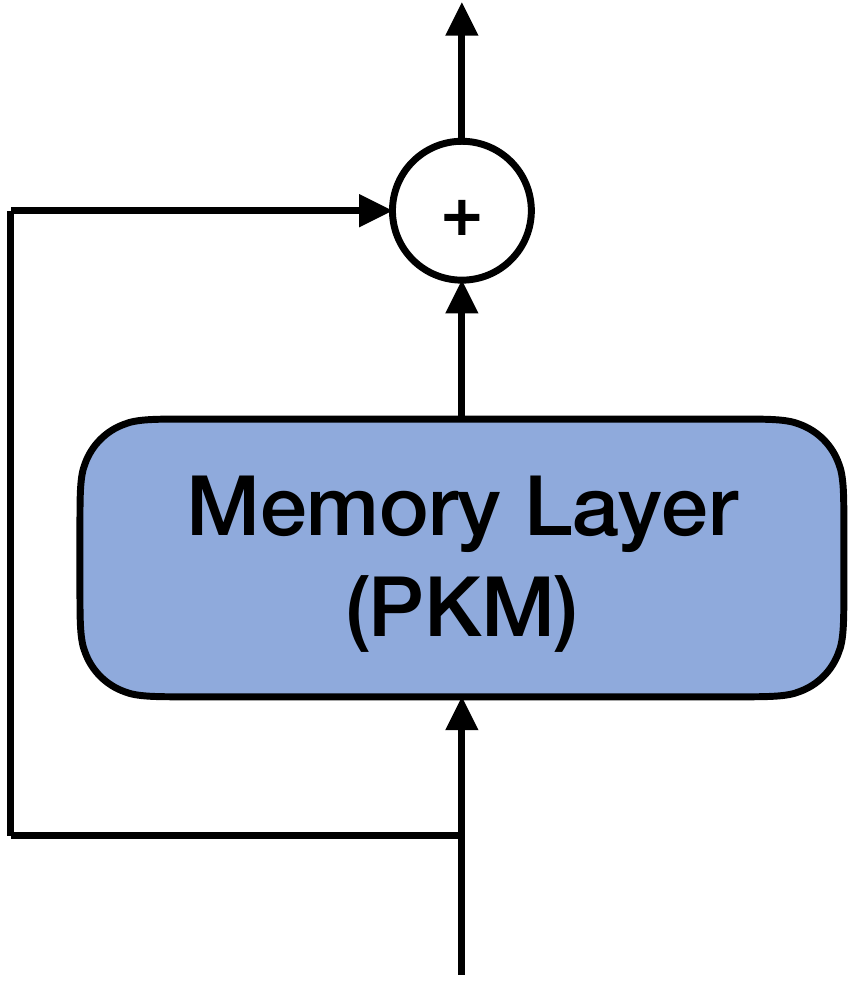}}
    \:\:\:\:
    \subfloat[\resm~layer]{\includegraphics[width=0.46\textwidth]{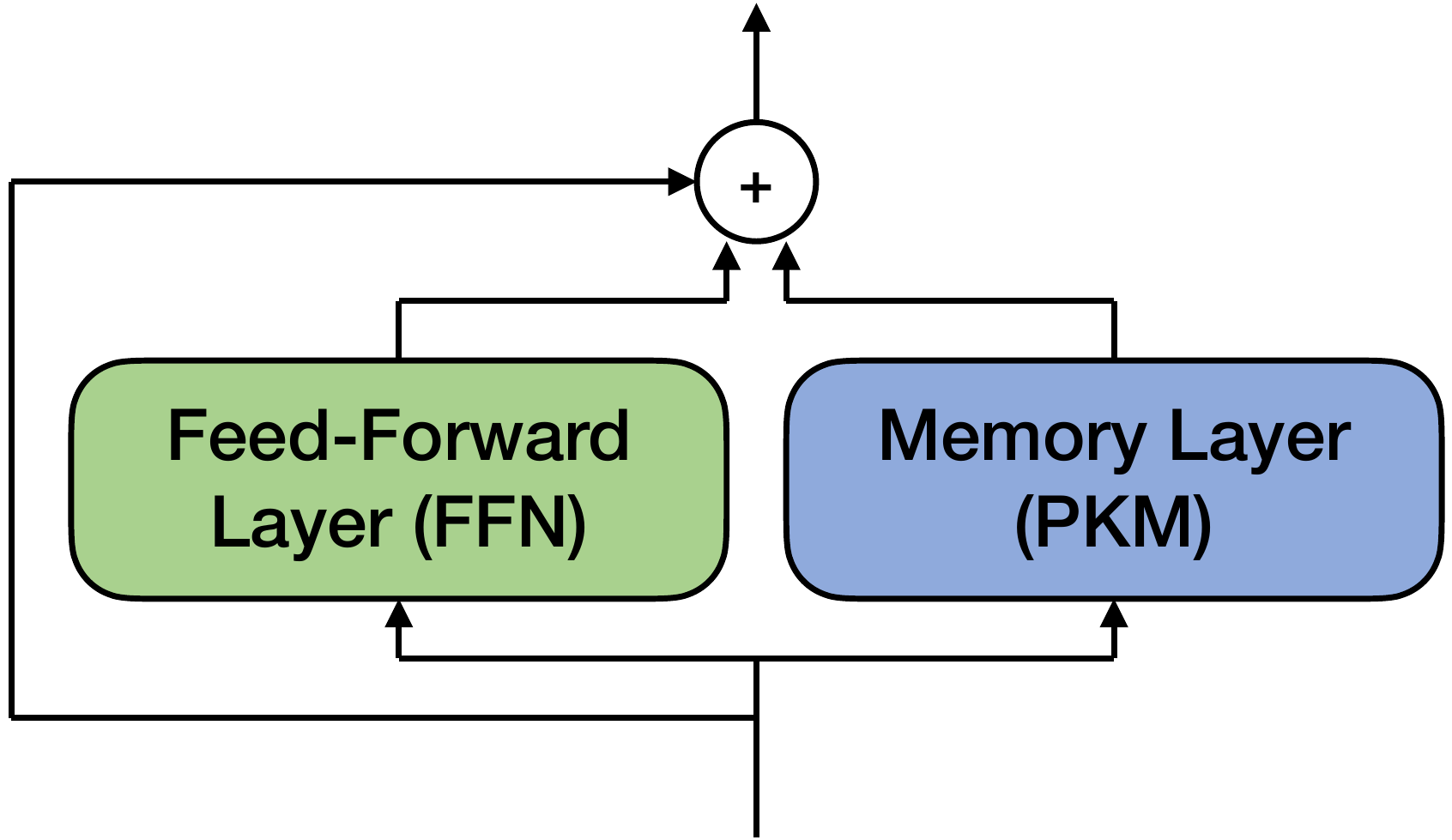}} \\
    \caption{
    \label{fig:residual_memory_layer}
    Illustration of the layers for the comparison. 
    (a) displays feed-forward layer (FFN) in vanilla Transformer architecture \cite{vaswani2017attention}.
    (b) is the original version of product-key memory (PKM) layer \cite{lample2019large} that replaces FFN.
    (c) is our proposed \resm~layer.
    Instead of replacing FFN to PKM, \resm~adds PKM in addition to FFN as a residual connection.
    }
\end{figure*}

%% file: 05_experiments.tex
\input{tabs/pretraining}

\section{Experiment Setup}
\label{sec:expr-setting}

\subsection{Product Key Memory}

Our implementation is based on HuggingFace's \textit{Transformers} library\footnote{\href{https://github.com/huggingface/transformers}{https://github.com/huggingface/transformers}} \cite{wolf2019transformers}, and the PKM part is borrowed from the XLM repository.\footnote{\href{https://github.com/facebookresearch/XLM}{https://github.com/facebookresearch/XLM}}
We add two memory layers in the intermediate layers at regular intervals: i.e., \{4,8\} in 12 layer models, and \{2,4\} in 6 layer models.
We will explore the effect of changing the number of the position of memory layers in the future.
We use 512\textsuperscript{2} ($\approx$ 262k) memory slots with 4 memory heads and select 32 keys per head for each memory layer for all experiments.
We set the dimension of key vectors and value vectors to 256 and 768, respectively.
We use query batch normalization to increase key coverage during training.
We measure the top-1 memory usage and the KL divergence to measure how much the model effectively uses memory capacity.

\subsection{Pretraining}

We use 12 layer \bb{} models with and without PKM.
For pretraining, we use English Wikipedia and BookCorpus \cite{zhu2015aligning} as a training corpus like BERT \cite{devlin2018bert}, in total 17GB.
We use the same vocabulary and tokenizer with \citet{devlin2018bert}.
We train models with batch size of 1024 sequences for 500,000 steps.
We use Adam optimizer \cite{kingma2014adam} with learning rate of 1e-4 and linear warmup scheduler over the first 10,000 steps.
The memory values are learned with a sparse update of learning rate 1e-3, following \citet{lample2019large}.
With half-precision training\footnote{\href{https://github.com/NVIDIA/apex}{https://github.com/NVIDIA/apex}} on 32 NVIDIA V100 GPUs, pretraining took 2.8 days without PKM and 5.1 days with PKM (or with ResM).

To evaluate pretrained models themselves, we measure the perplexity of masked language modeling on the test set of WikiText-2, WikiText-103, and PG-19 \cite{rae2019compressive}.  %
Since the pretraining corpus covers WikiText-2 and WikiText-103, perplexity on them is a proxy to the training perplexity.
Meanwhile, because the PG-19 dataset came from different sources, perplexity on PG-19 can be regarded as the test perplexity.

\subsection{Finetuning}

\input{tabs/finetuning_details}

For fine-tuning, we use SQuAD 1.1 \cite{rajpurkar2016squad} and GLUE \cite{wang2018glue} benchmark as downstream tasks.
Following other PLM literature, including RoBERTa (Liu et al., 2019), we report dev set results instead of the test set to compare our variants.
We report a median of 5 runs with different random seeds for each fine-tuning task.
We measure exact match (EM) and F1 scores on SQuAD 1.1. 
For QQP, which is the binary classification task, the F1 score is used for the GLUE leaderboard.
However, we use the accuracy as the metric for development set because the F1 score varies a lot depending on random seeds.
Finetuning details appear in Table \ref{tab:finetuning_details}.

%% file: tabs/pretraining.tex
\begin{table*}[t]
    \centering
    \footnotesize
    
    \begin{tabular*}{0.75\textwidth}{+l@{\extracolsep{\fill}}=c=c=c=c=c=c=c}
    \toprule
    \multicolumn{1}{c}{\multirow{3}*{Model}} & 
    \multicolumn{3}{c}{Memory} & 
    \multicolumn{3}{c}{MLM} \\
    \rowstyle{\scriptsize}
    & \tmemu & \klu & \klw & WT-2 & WT-103 & PG-19 \\
    \rowstyle{\scriptsize}
    & (4L/8L) (\%) & (4L/8L) & (4L/8L) & (ppl) & (ppl) & (ppl) \\
    \midrule
    (a) \bb$^{\dagger}$ & - & - & - & 3.49 & 3.86 & 6.18 \\
    (b) \:\: +500k steps  & - & - & - & 3.40 & 3.72 & 5.88  \\
    \midrule
    (c) \:\: +PKM         &  2.2/84.1 & 1.62/0.89 & 1.99/1.13 & 3.26 & 3.39 & 5.53 \\
    (d) \:\: +ResM        & 75.0/81.0 & 1.50/0.71 & 1.80/0.92 & 3.26 & 3.36 & 5.45\\
    (e) \:\: +Init +PKM   & 97.4/95.7 & 0.53/0.69 & 0.68/0.88 & 3.14 & 3.26 & 5.22 \\
    (f) \:\: +Init +ResM  & \textbf{98.2}/\textbf{97.3} & \textbf{0.45}/\textbf{0.46} & \textbf{0.58}/\textbf{0.60} & \textbf{3.10} & \textbf{3.20} & \textbf{5.14} \\
    \bottomrule
    \end{tabular*}

    \caption{
    \label{tab:pretraining}
    Experimental results of pre-training PKM-augmented PLMs.
    Because standard memory usage is almost 100\%, we omit it in the table. 
    Top-1 memory usage and KL divergence are calculated at the 4th and 8th layers.
    $^{\dagger}$: we pre-train BERT\textsubscript{BASE} by ourself.
    }
\end{table*}

%% file: tabs/finetuning_details.tex
\begin{table}[!t]
    \centering
    \setlength{\tabcolsep}{0.5pt}
    \footnotesize
    \begin{tabular*}{0.48\textwidth}{l@{\extracolsep{\fill}}rrrrrrr}
    \toprule
    \multicolumn{1}{c}{Dataset} & 
    \multicolumn{1}{c}{lr} & 
    \multicolumn{1}{c}{bsz} & 
    \multicolumn{1}{c}{\# epoch} & 
    \multicolumn{1}{c}{\makecell{warmup \\ ratio}} &
    \multicolumn{1}{c}{\makecell{weight \\ decay}} &
    \multicolumn{1}{c}{\makecell{max seq \\ length}} \\
    \midrule
    SQuAD 1.1  & 5e-5 & 32 & 3 & 0.06 & 0.01 & 384 \\
    GLUE       & 2e-5 & 32 & 10 & 0.06 & 0.1 & 128  \\
    \bottomrule
    \end{tabular*}
    \caption{
    \label{tab:finetuning_details}
    Fine-tuning hyper-parameters for downstream tasks, SQuAD 1.1 and GLUE. We use 128 doc stride for SQuAD 1.1 dataset.
    }
\end{table}

%% file: 06_pretraining_results.tex
\section{Pretraining Results}
\label{sec:expr-pretraining}

Table \ref{tab:pretraining} shows the experimental results of pretraining.
We compare models with/without the initialization and PKM vs. ResM.
We use \bb{} architecture of 12 transformer layers without next sentence prediction following \citet{liu2019roberta} for our pretraining experiments.
For the fair comparison between \bb{} and PKM-augmented-\bb{} after the initialization, we train \bb{} with longer steps, but the improvement was marginal.
    
\medskip \noindent
\textbf{Memory Utilization}
Surprisingly, the top-1 memory usage of the PKM-augmented PLM at the 4th layer is about 2\%, which is remarkably low, though top-32 memory usage at this layer is almost 100\%.
In other words, the model does not take advantage of the lower memory layer effectively.

With a residual connection, the top-1 memory usage of all layers become reasonably high.
Similar to \citet{he2016deep}, the residual connection helps to learn deep networks with memory, resulting in improved accuracy.
Moreover, with the initialization from pretrained weights, top-1 memory usage is more than 95\%.
With the initialization and ResM, top-1 memory usage increases, and KL divergence decreases significantly, implying better exploitation of the memory layers.
It becomes possible by preventing memory parameters not to suffer from the catastrophic drift.

\input{figs/staleness}

We check when each memory slot is used at last among saved checkpoints.
Then, we count the number of slots depending on the last used checkpoint.
Figure \ref{fig:staleness} indirectly indicates how many memory slots are kept not selected as top-1.
This figure provides evidence that a model with the initialization and residual memory prevents staleness compared to a model with plain PKM.

\input{figs/training_curve}

\input{tabs/finetuning}

\medskip \noindent
\textbf{Masked Language Modeling}
Augmenting large PKM always improves masked language modeling compared to a model without memory.
Figure \ref{fig:training_curve} shows the training curve of the models after the initialization.
It proves that the residual connection prevents a deviation of the PKM at the beginning (bigger initial perplexity) even with the initialization from the pretrained weight.
Although they are converged to a similar perplexity after very long training steps, the initial perplexity of PKM is much bigger than that of \resm.
In sum, both the initialization from pretrained PLM and the residual memory layer are beneficial for PLM with a memory to perform better in masked language modeling.

%% file: figs/staleness.tex
\begin{figure}[t!]
    \centering 
    \subfloat[4th Layer]{\includegraphics[width=0.42\textwidth]{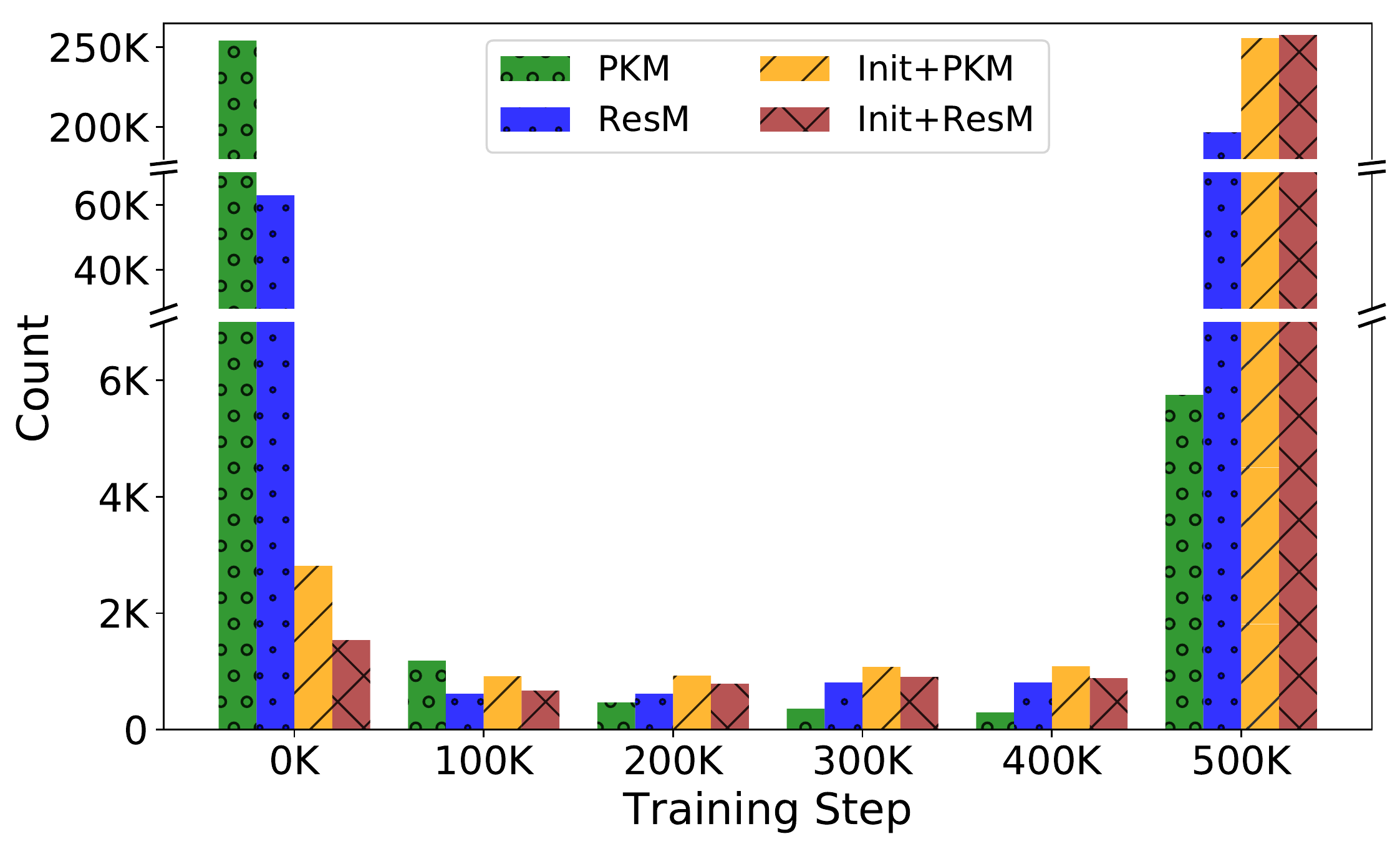}}
    
    \subfloat[8th Layer]{\includegraphics[width=0.42\textwidth]{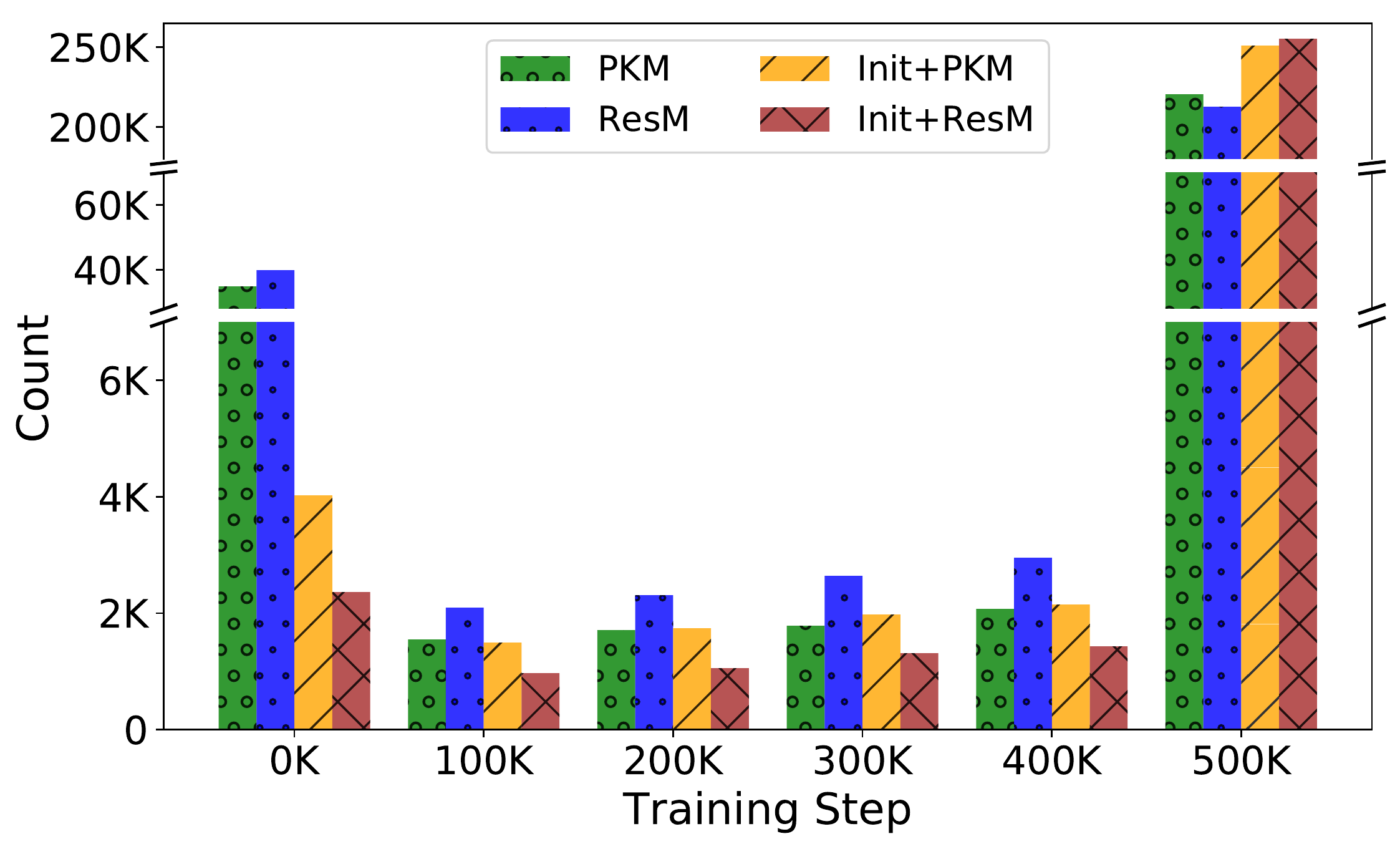}}
    \caption{
    \label{fig:staleness}
    Histogram for staleness evaluation of PKM-augmented PLMs.
    We save model checkpoints every 100k step during the entire 500k pre-training steps.
    This histogram illustrates how many memory slots are used at last for each saved checkpoint.
    For example, if a key is used at 200k model checkpoint and never used after that, then it is likely to keep its state as stale after 200k.
    Because the total number of memory slots is fixed to 512\textsuperscript{2}, the model having boxes toward the right in the graph is better.
    }
\end{figure}

%% file: figs/training_curve.tex
\begin{figure}[t]
    \centering 
    \includegraphics[width=0.44\textwidth]{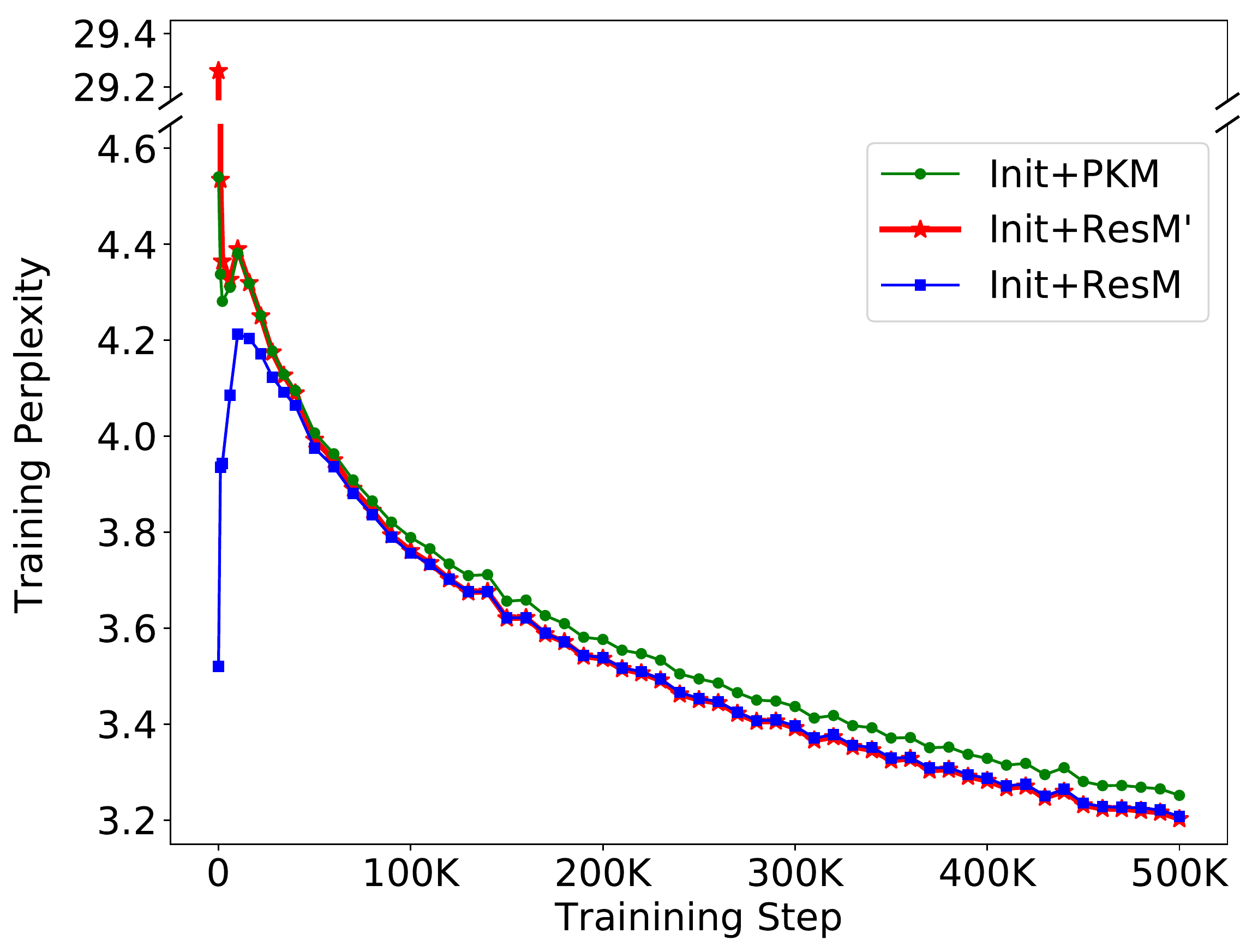}
    \caption
    {
    \label{fig:training_curve}
    Training curves of MLM perplexity versus training steps during the pre-training of PKM-augmented MLMs.
    Y-axis is zoomed in the low perplexity region.
    Initialized from the pre-trained BERT, PKM (green) replaces FFN to PKM, ResM (blue) add PKM as a residual connection.
    ResM' (red) is the same with ResM but randomly re-initialize FFN. 
    } 
\end{figure}

%% file: tabs/finetuning.tex
\begin{table*}[t]
    \centering
    \footnotesize
    
    \begin{tabular*}{0.8\textwidth}{+l@{\extracolsep{\fill}}=c=c=c=c=c=c=c}
    \toprule
    \multicolumn{1}{c}{\multirow{3}*{Model}} & 
    QA & 
    \multicolumn{6}{c}{GLUE} \\
    \rowstyle{\scriptsize}
    & SQuAD 1.1 & MNLI{\tiny -(m/mm)} & QQP & QNLI & SST-2 & CoLA & {\scriptsize \textbf{Avg}} \\
    \rowstyle{\scriptsize}
    & (EM/F1) & (Acc) & (Acc) & (Acc) & (Acc) & (Matt) & - \\
    \midrule
    (a) \bb$^{\dagger}$     & 82.7/89.8 & 84.3/84.5 & 91.0 & 89.3 & 92.8 & 60.8 & 83.8  \\
    (b) \:\: +500k steps    & 83.3/90.1 & 84.8/84.9 & 91.2 & 89.2 & 92.4 & 61.4 & 84.0 \\  
    \midrule
    (c) \:\: +PKM           & 81.9/89.1 & 84.4/85.0 & 91.1 & 89.0 & 93.6 & 59.7 & 83.8 \\
    (d) \:\: +ResM          & 81.5/89.4 & 84.6/84.8 & 91.0 & 88.2 & 93.2 & 62.8 & 84.1 \\
    (e) \:\: +Init +PKM     & 83.8/90.6 & 85.8/85.6 & 91.2 & 90.0 & 93.6 & 63.6 & 85.0 \\
    (f) \:\: +Init +ResM    & \textbf{83.9}/\textbf{90.8} & \textbf{86.0}/\textbf{85.8} & \textbf{91.4} & \textbf{90.4} & \textbf{94.0} & \textbf{64.1} & \textbf{85.3} \\
    \midrule
    \midrule
    (g) \bb$^{\star}$       & 81.1/88.5 & 83.9/84.4 & 91.0 & 88.4 & 92.9 & 59.8 & 83.4 \\
    (h) \bl$^{\star}$       & 83.3/90.6 & 86.2/86.1 & 91.4 & 90.4 & 93.8 & 64.1 & 85.3 \\
    \bottomrule
    \end{tabular*}
    
    \caption{
    \label{tab:finetuning}
    Experimental results of fine-tunining PKM-augmented PLMs.
    Model (a)-(f) are the same one from Table \ref{tab:pretraining}.
    \textsuperscript{$\star$}: we borrow pretrained weights of \bb~and \bl~from \cite{devlin2018bert}.
    We fine-tune these models on SQuAD 1.1 \cite{rajpurkar2016squad} and GLUE tasks \cite{wang2018glue}.
    }
\end{table*}

%% file: 07_finetuning_results.tex
\section{Finetuning Results}

Table \ref{tab:finetuning} shows the experimental results of finetuning using our pretrained models.

\input{tabs/memory_fix}

\medskip \noindent
\textbf{Downstream Performance}
Although large PKM helps masked language modeling, the downstream performance of several tasks with plain PKM is worse than the baseline without memory.
We think this is because the catastrophic drift problem is especially severe in the fine-tuning step.
Downstream dataset size and the number of training steps are too small to fit memory parameters accordingly.

Better memory utilization coming from the initialization and the residual connection also leads to better downstream accuracy in most of the datasets.
We report the fine-tuning results using the weights of pretrained \bl{} from \citet{devlin2018bert} in Table~\ref{tab:finetuning}.\footnote{
Unfortunately, we could not pretrain \bl, so we will prepare it after the submission.
In our pretraining experiments, we use almost same settings but larger batch size (256 vs 1024) than \citet{devlin2018bert}.
The difference between our pretrained \bb{} (a) and Google \bb{} (g) and the difference between our ResM-augmented \bb{} with the initialization (f) and Google \bl{} (h) are insignificant.
}
We believe that our best PKM-augmented-\bb{} would have comparable performance with \bl{} even after pretraining it by ourselves, while much faster as described in Table \ref{tab:speed}.

On the assumption that updating memory parameters sparsely using a limited number of data and training steps might be vulnerable to the catastrophic drift, we try to fix memory parameters during fine-tuning as in Table \ref{tab:memory_fix}.
However, it degrades the downstream performance.

\input{tabs/mem_finetuning}

\input{figs/class_difference}

\medskip \noindent
\textbf{Memory Utilization}
Table \ref{tab:mem_finetuning} shows the memory usage and KL divergence of fine-tuned PKM-augmented models.
Comparison of fine-tuned PKM-augmented models in terms of the memory usage has similar trends with that of pretraining.
The initialization and the residual memory improve memory usage, meaning better exploitation of model capacity for downstream tasks.
Especially in a large dataset like MNLI \cite{williams2017broad}, the memory usage of the fine-tuned model reaches to almost 100\% similar to pretrained models due to the sufficient training steps to update memory parameters.
On the other hand, interestingly, the initialization and the residual memory do not always reduce KL divergence.
We presume this because fine-tuning of classification tasks encourages input examples of the same class to be clustered into similar representations, so it requires to access similar patterns of memory slots while utilizing many of them.

To validate the assumption mentioned above, we check the difference in memory usage between positive examples and negative examples using SST-2 \cite{socher2013recursive} dataset, which is the binary classification tasks to predict the sentiment of a movie review.
To measure the difference, we calculate (1) KL divergence between two distributions (positive/negative) and (2) intersection over union (IOU), which is a widely used metric in object detection \cite{ren2015faster} on the top-1 memory usage.
We calculate IOU as $\sum_i min(t_i^+, t_i^-) / \sum_i max(t_i^+, t_i^-)$, where $t_i^+$ and $t_i^-$ is a top-1 usage at memory position $i$ for positive examples and negative examples, respectively.
As illustrated in Figure \ref{fig:class_difference}, our best PKM-augmented model shows much higher KL and lower IOU in every layer than the plain PKM-augmented model, implying better discriminative ability. 

\input{tabs/small}

\medskip \noindent
\textbf{Other Pretrained Models}
We release the code and pretrained weights to encourage researchers and practitioners to easily utilize and reproduce our work, allowing the application to different model sizes and other backbone architectures. 
In particular, we employ our methods to DistilBERT model \citep{sanh2019distilbert}, which is a 6-layer transformer model trained by knowledge distillation \citep{hinton2015distilling} from \bb.
Similarly, it obtains accuracy comparable to \bb{} as shown in Table \ref{tab:small}.\footnote{
Like Table \ref{tab:finetuning}, we borrow \db{} and \bb{} of public weights, so their training setting does not match.
We use a batch size of 2048 for pretraining ResM-augmented \db{} initialized from the weights of \db{} (c.f. batch size of \db{} and \bb{} in their training is 4096 and 256, respectively).
}
Moreover, we believe our approaches could also be helpful to any other task.

\medskip \noindent
\textbf{PKM vs. ResM}
One might argue that the gap between the PKM model and the ResM model might be attributed to the difference in model size.
We claim that the impact of the architectural difference between PKM and ResM is more than from more parameters.
ResM achieves better memory utilization, resulting in a better final performance. 
0.3 higher average GLUE score with only 9M more parameters (smaller than 2\% of the entire model) is significant considering that BERT-Large achieves a 1.9 higher average GLUE score with 230M more parameters than BERT-Base ($\frac{0.3}{9} \gg \frac{1.9}{230}$).

%% file: tabs/memory_fix.tex
\begin{table}[t]
    \centering
    \setlength{\tabcolsep}{1pt}
    \footnotesize
    
    \begin{tabular*}{0.48\textwidth}{+l@{\extracolsep{\fill}}=c=c=c=c=c}
    \toprule
    \multicolumn{1}{c}{\multirow{3}*{Model}} & 
    \multirow{3}*{\makecell{Memory \\ Update}} & 
    QA & 
    \multicolumn{3}{c}{GLUE} \\
    \rowstyle{\scriptsize}
    & & SQuAD 1.1 & {\scriptsize MNLI}{\tiny -m} & {\scriptsize SST-2} & {\scriptsize CoLA} \\
    \rowstyle{\scriptsize}
    &  & (EM/F1) & (Acc) & (Acc) & (Matt) \\
    \midrule
    \multirow{2}*{(c) +PKM} & Y & 81.9/89.1 & 84.4 & 93.6 & 59.7  \\
     & N & 82.0/89.0 & 84.1 & 93.0 & 56.5   \\
    \midrule[0.1pt]
    \multirow{2}*{(d) +ResM} & Y & 81.5/89.4 & 84.6 & 93.2 & 62.8 \\
     & N & 82.2/89.5 & 84.3 & 92.7 & 59.9  \\
    \midrule[0.1pt]
    \multirow{2}*{(e) +Init +PKM} & Y & 83.8/90.6 & 85.8 & 93.6 & 63.6 \\
     & N & 83.7/90.4 & 85.5 & 93.3 & 58.8 \\
    \midrule[0.1pt]
    \multirow{2}*{(f) +Init +ResM} & Y & 83.9/90.8 & 86.0 & 94.0 & 64.1 \\
     & N & 84.2/90.8 & 85.8 & 93.3 & 61.6 \\
    \bottomrule
    \end{tabular*}
    
    \caption{
    \label{tab:memory_fix}
    Ablation study on fixing memory parameters during fine-tuning.
    }
\end{table}

%% file: tabs/mem_finetuning.tex
\begin{table*}[t!]
    \centering
    \setlength{\tabcolsep}{0.5pt}
    \footnotesize
    \begin{tabular*}{0.8\textwidth}{+l@{\extracolsep{\fill}}=c=c=c=c=c=c=c=c=c=c=c=c=c=c=c=c=c}
    \toprule
    \multicolumn{1}{c}{\multirow{3}*{Model}} & 
    \multirow{3}*{\makecell{Memory \\ Position}} & 
    \multicolumn{4}{c}{MNLI{\scriptsize{-m}}} & 
    \multicolumn{4}{c}{SST-2} & 
    \multicolumn{4}{c}{CoLA}
    \\
    \rowstyle{\scriptsize}
    & & \memu & \tmemu & \klu & \klw 
    & \memu & \tmemu & \klu & \klw 
    & \memu & \tmemu & \klu & \klw  \\
    \rowstyle{\scriptsize}
    &  & (\%) & (\%) &  &  
    & (\%) & (\%) &  &  
    & (\%) & (\%) &  &  \\
    \midrule
    \multirow{2}*{(c) +PKM}         & 4 &  99.4 &  1.2 & 2.14 & 2.36 & 79.0 &  0.6 & 3.23 & 3.49 & 60.5 &  0.4 & 4.62 & 4.89  \\
                                    & 8 &  99.7 & 68.2 & 2.30 & 2.47 & 83.7 & 35.8 & 2.59 & 2.76 & 61.8 & 22.2 & 5.51 & 5.67  \\
    \midrule[0.1pt]
    \multirow{2}*{(d) +ResM}        & 4 &  98.9 & 64.5 & 2.46 & 2.71 & 79.3 & 35.6 & 3.84 & 4.08 & 61.5 & 24.1 & 4.01 & 4.20  \\
                                    & 8 &  99.9 & 66.7 & 1.87 & 2.02 & 84.6 & 32.5 & \textbf{2.34} & \textbf{2.51} & 73.3 & 22.7 & \textbf{2.05} & \textbf{2.21}  \\
    \midrule[0.1pt]
    \multirow{2}*{(e) +Init +PKM}   & 4 & \textbf{100.0} & 81.8 & 1.33 & 1.46 & 91.2 & 42.3 & 3.52 & 3.76 & 72.7 & 26.3 & 4.11 & 4.29  \\
                                    & 8 &  99.9 & 78.5 & 1.76 & 1.95 & 86.3 & 35.8 & 2.81 & 3.05 & 65.5 & 21.9 & 4.72 & 4.93  \\
    \midrule[0.1pt]
    \multirow{2}*{(f) +Init +ResM}  & 4 & \textbf{100.0} & \textbf{85.6} & \textbf{0.94} & \textbf{1.06} & \textbf{92.0} & \textbf{42.8} & \textbf{2.98} & \textbf{3.18} & \textbf{75.5} & \textbf{28.6} & \textbf{3.99} & \textbf{4.15}  \\
                                    & 8 & \textbf{100.0} & \textbf{85.6} & \textbf{1.52} & \textbf{1.66} & \textbf{89.9} & \textbf{41.6} & 2.39 & 2.63 & \textbf{73.6} & \textbf{27.4} & 3.88 & 4.06  \\
    \bottomrule
    \end{tabular*}
    \caption{
    \label{tab:mem_finetuning}
    Memory utilization of PKM-augmented models after fine-tuning.
    We measure memory utilization metrics (\memu, \tmemu, \klu, and \klw) at 4th and 8th layer after fine-tuning using MNLI-m \cite{williams2017broad}, SST-2 \cite{socher2013recursive}, and CoLA \cite{warstadt2019neural} datasets as an example.
    We use the same fine-tuned models that appeared in Table \ref{tab:pretraining}.
    }
\end{table*}

%% file: figs/class_difference.tex
\begin{figure}[t]
    \centering 

    \includegraphics[width=0.48\textwidth]{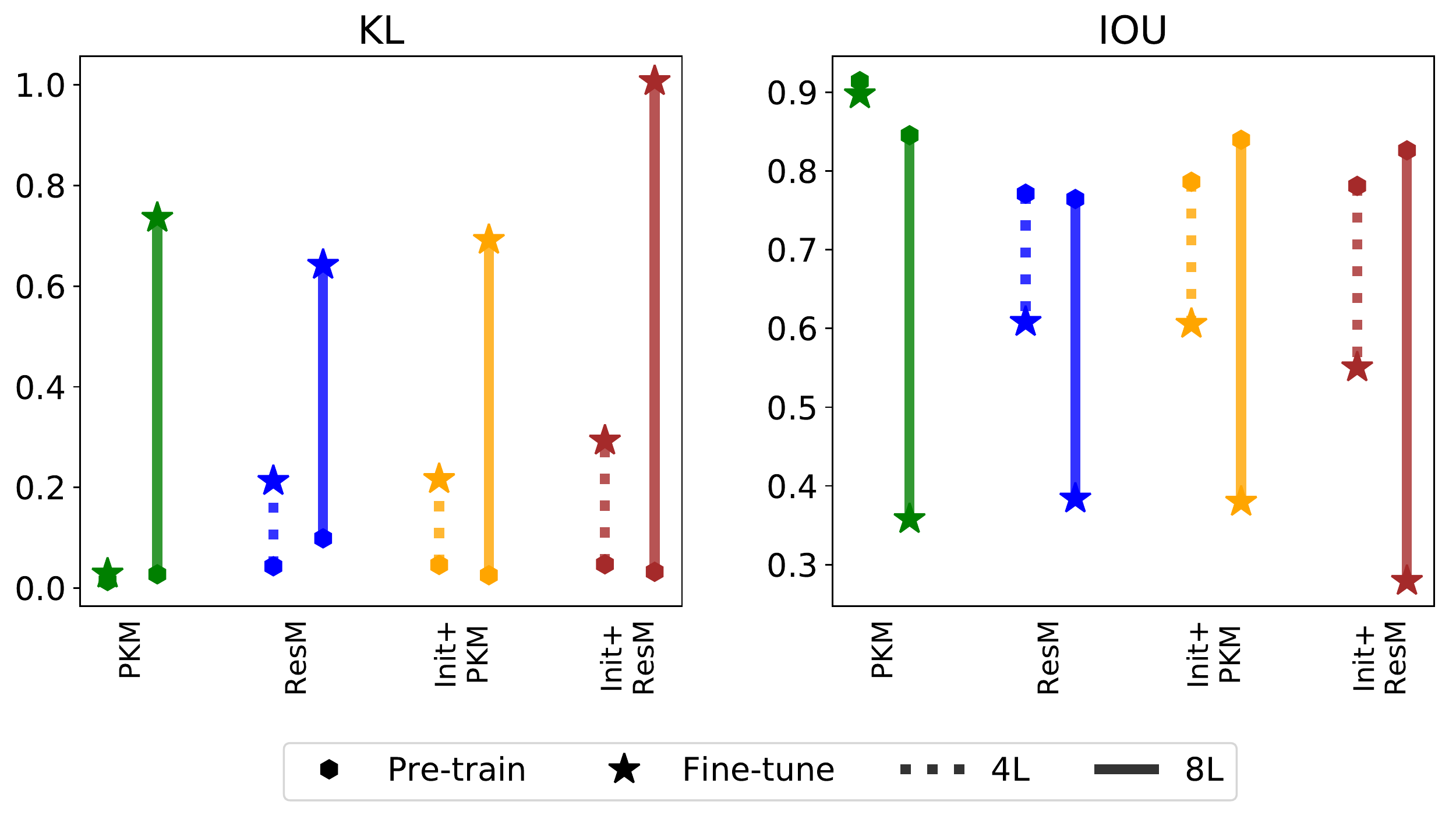}

    \caption
    {
    \label{fig:class_difference}
    Difference in memory usage between positive examples and negative examples in SST-2 \cite{socher2013recursive}.
    KL divergence (left) and IOU (right) between two distributions (positive vs. negative) are visualized.
    We measure those values from weights pre-trained without fine-tuning and after fine-tuning.
    } 
\end{figure}

%% file: tabs/small.tex
\begin{table}[!t]
    \centering
    \setlength{\tabcolsep}{1pt}
    \footnotesize
    
    \begin{tabular*}{0.48\textwidth}{+l@{\extracolsep{\fill}}=r=c=c=c}
    \toprule
    \multicolumn{1}{c}{\multirow{3}*{Model}} & 
    \multicolumn{1}{c}{MLM} & 
    QA & 
    \multicolumn{2}{c}{GLUE}  \\
    \rowstyle{\scriptsize}
     & PG-19 & SQuAD 1.1 & MNLI{\scriptsize -m} & SST-2 \\
    \rowstyle{\scriptsize}
     & (ppl) & (EM/F1) & (Acc) & (Acc) \\
    \midrule
    \db\textsuperscript{$\star$} & 20.61 & 77.4/85.7 & 82.0 & 91.6 \\
    \:\: +Init +ResM             &  5.75 & 80.4/88.3 & 84.1 & 93.3 \\
    \midrule
    \midrule
    \bb\textsuperscript{$\star$} & 11.82 & 81.1/88.5 & 83.9 & 92.9  \\
    \bottomrule
    \end{tabular*}
    
    \caption{
    \label{tab:small}
    Experimental results on DistilBERT with and without our method.
    We add results of \bb~for comparison.
    \textsuperscript{$\star$} means our reproduced results using model weights from \citet{sanh2019distilbert} and \citet{devlin2018bert}.
    The initial model weights of DistilBERT is from the part of \bb.
    }
\end{table}

%% file: 08_conclusion_and_future_work.tex
\section{Conclusion and Future Work}

This work starts from unexpected results that directly applying PKM to PLMs does not work well in downstream tasks, contrary to \citep{lample2019large}.
In this paper, we successfully augment PKM to PLMs with two ingredients, weight initialization and residual connection, based on the observation of memory utilization and catastrophic drift during the training.
Consequently, we encourage to utilize memory architecture such as PKM for PLMs in practical use.

Although our approach mitigates the catastrophic drift problem somehow, we leave further study on it during both pretraining and finetuning as future work.
One possible solution is to regularize a PKM memory by a structured dropout on the memory keys like DropHead \cite{zhou2020scheduled}.
It would also help to prune unnecessary memory slots on-demand during the inference time.

%% file: 09_acknowledgements.tex
\section*{Acknowledgments}
The authors would like to thank Clova AI members and the anonymous reviewers for their constructive feedback.
Primarily, LaRva (Language Representation by Clova) team members, including Sungdong Kim, Dongjun Lee, and Minjeong Kim, provided their full support to this work happens.
Moreover, Guillaume Lample shared his experience and tips for our preliminary experiments regarding product key memory.
Lastly, GK greatly appreciates Kyunghyun Cho, Minjoon Seo, Jinhyuk Lee, and Jung-Woo Ha for proofreading this manuscript thoroughly.
We use Naver Smart Machine Learning \cite{sung2017nsml,kim2018nsml} platform for the experiments.

%% file: ms.bbl
\begin{thebibliography}{33}
\expandafter\ifx\csname natexlab\endcsname\relax\def\natexlab#1{#1}\fi

\bibitem[{Ba et~al.(2016)Ba, Kiros, and Hinton}]{ba2016layer}
Jimmy~Lei Ba, Jamie~Ryan Kiros, and Geoffrey~E Hinton. 2016.
\newblock Layer normalization.
\newblock \emph{arXiv preprint arXiv:1607.06450}.

\bibitem[{Brown et~al.(2020)Brown, Mann, Ryder, Subbiah, Kaplan, Dhariwal,
  Neelakantan, Shyam, Sastry, Askell et~al.}]{brown2020language}
Tom~B Brown, Benjamin Mann, Nick Ryder, Melanie Subbiah, Jared Kaplan, Prafulla
  Dhariwal, Arvind Neelakantan, Pranav Shyam, Girish Sastry, Amanda Askell,
  et~al. 2020.
\newblock Language models are few-shot learners.
\newblock \emph{arXiv preprint arXiv:2005.14165}.

\bibitem[{Chandar et~al.(2016)Chandar, Ahn, Larochelle, Vincent, Tesauro, and
  Bengio}]{chandar2016hierarchical}
Sarath Chandar, Sungjin Ahn, Hugo Larochelle, Pascal Vincent, Gerald Tesauro,
  and Yoshua Bengio. 2016.
\newblock Hierarchical memory networks.
\newblock \emph{arXiv preprint arXiv:1605.07427}.

\bibitem[{Devlin et~al.(2018)Devlin, Chang, Lee, and
  Toutanova}]{devlin2018bert}
Jacob Devlin, Ming-Wei Chang, Kenton Lee, and Kristina Toutanova. 2018.
\newblock Bert: Pre-training of deep bidirectional transformers for language
  understanding.
\newblock \emph{arXiv preprint arXiv:1810.04805}.

\bibitem[{F{\'e}vry et~al.(2020)F{\'e}vry, Soares, FitzGerald, Choi, and
  Kwiatkowski}]{fevry2020entities}
Thibault F{\'e}vry, Livio~Baldini Soares, Nicholas FitzGerald, Eunsol Choi, and
  Tom Kwiatkowski. 2020.
\newblock Entities as experts: Sparse memory access with entity supervision.
\newblock \emph{arXiv preprint arXiv:2004.07202}.

\bibitem[{Guu et~al.(2020)Guu, Lee, Tung, Pasupat, and Chang}]{guu2020realm}
Kelvin Guu, Kenton Lee, Zora Tung, Panupong Pasupat, and Ming-Wei Chang. 2020.
\newblock Realm: Retrieval-augmented language model pre-training.
\newblock \emph{arXiv preprint arXiv:2002.08909}.

\bibitem[{He et~al.(2016)He, Zhang, Ren, and Sun}]{he2016deep}
Kaiming He, Xiangyu Zhang, Shaoqing Ren, and Jian Sun. 2016.
\newblock Deep residual learning for image recognition.
\newblock In \emph{Proceedings of the IEEE conference on computer vision and
  pattern recognition}, pages 770--778.

\bibitem[{Hinton et~al.(2015)Hinton, Vinyals, and Dean}]{hinton2015distilling}
Geoffrey Hinton, Oriol Vinyals, and Jeff Dean. 2015.
\newblock Distilling the knowledge in a neural network.
\newblock \emph{arXiv preprint arXiv:1503.02531}.

\bibitem[{Khandelwal et~al.(2019)Khandelwal, Levy, Jurafsky, Zettlemoyer, and
  Lewis}]{khandelwal2019generalization}
Urvashi Khandelwal, Omer Levy, Dan Jurafsky, Luke Zettlemoyer, and Mike Lewis.
  2019.
\newblock Generalization through memorization: Nearest neighbor language
  models.
\newblock \emph{arXiv preprint arXiv:1911.00172}.

\bibitem[{Kim et~al.(2018)Kim, Kim, Seo, Kim, Park, Park, Jo, Kim, Yang, Kim
  et~al.}]{kim2018nsml}
Hanjoo Kim, Minkyu Kim, Dongjoo Seo, Jinwoong Kim, Heungseok Park, Soeun Park,
  Hyunwoo Jo, KyungHyun Kim, Youngil Yang, Youngkwan Kim, et~al. 2018.
\newblock Nsml: Meet the mlaas platform with a real-world case study.
\newblock \emph{arXiv preprint arXiv:1810.09957}.

\bibitem[{Kingma and Ba(2014)}]{kingma2014adam}
Diederik~P Kingma and Jimmy Ba. 2014.
\newblock Adam: A method for stochastic optimization.
\newblock \emph{arXiv preprint arXiv:1412.6980}.

\bibitem[{Lample et~al.(2019)Lample, Sablayrolles, Ranzato, Denoyer, and
  J{\'e}gou}]{lample2019large}
Guillaume Lample, Alexandre Sablayrolles, Marc'Aurelio Ranzato, Ludovic
  Denoyer, and Herv{\'e} J{\'e}gou. 2019.
\newblock Large memory layers with product keys.
\newblock In \emph{Advances in Neural Information Processing Systems}, pages
  8546--8557.

\bibitem[{Liu et~al.(2019)Liu, Ott, Goyal, Du, Joshi, Chen, Levy, Lewis,
  Zettlemoyer, and Stoyanov}]{liu2019roberta}
Yinhan Liu, Myle Ott, Naman Goyal, Jingfei Du, Mandar Joshi, Danqi Chen, Omer
  Levy, Mike Lewis, Luke Zettlemoyer, and Veselin Stoyanov. 2019.
\newblock Roberta: A robustly optimized bert pretraining approach.
\newblock \emph{arXiv preprint arXiv:1907.11692}.

\bibitem[{Rae et~al.(2016)Rae, Hunt, Danihelka, Harley, Senior, Wayne, Graves,
  and Lillicrap}]{rae2016scaling}
Jack Rae, Jonathan~J Hunt, Ivo Danihelka, Timothy Harley, Andrew~W Senior,
  Gregory Wayne, Alex Graves, and Timothy Lillicrap. 2016.
\newblock Scaling memory-augmented neural networks with sparse reads and
  writes.
\newblock In \emph{Advances in Neural Information Processing Systems}, pages
  3621--3629.

\bibitem[{Rae et~al.(2019)Rae, Potapenko, Jayakumar, and
  Lillicrap}]{rae2019compressive}
Jack~W Rae, Anna Potapenko, Siddhant~M Jayakumar, and Timothy~P Lillicrap.
  2019.
\newblock Compressive transformers for long-range sequence modelling.
\newblock \emph{arXiv preprint arXiv:1911.05507}.

\bibitem[{Raffel et~al.(2019)Raffel, Shazeer, Roberts, Lee, Narang, Matena,
  Zhou, Li, and Liu}]{raffel2019exploring}
Colin Raffel, Noam Shazeer, Adam Roberts, Katherine Lee, Sharan Narang, Michael
  Matena, Yanqi Zhou, Wei Li, and Peter~J Liu. 2019.
\newblock Exploring the limits of transfer learning with a unified text-to-text
  transformer.
\newblock \emph{arXiv preprint arXiv:1910.10683}.

\bibitem[{Rajpurkar et~al.(2016)Rajpurkar, Zhang, Lopyrev, and
  Liang}]{rajpurkar2016squad}
Pranav Rajpurkar, Jian Zhang, Konstantin Lopyrev, and Percy Liang. 2016.
\newblock Squad: 100,000+ questions for machine comprehension of text.
\newblock \emph{arXiv preprint arXiv:1606.05250}.

\bibitem[{Ren et~al.(2015)Ren, He, Girshick, and Sun}]{ren2015faster}
Shaoqing Ren, Kaiming He, Ross Girshick, and Jian Sun. 2015.
\newblock Faster r-cnn: Towards real-time object detection with region proposal
  networks.
\newblock In \emph{Advances in neural information processing systems}, pages
  91--99.

\bibitem[{Sanh et~al.(2019)Sanh, Debut, Chaumond, and
  Wolf}]{sanh2019distilbert}
Victor Sanh, Lysandre Debut, Julien Chaumond, and Thomas Wolf. 2019.
\newblock Distilbert, a distilled version of bert: smaller, faster, cheaper and
  lighter.
\newblock \emph{arXiv preprint arXiv:1910.01108}.

\bibitem[{Shoeybi et~al.(2019)Shoeybi, Patwary, Puri, LeGresley, Casper, and
  Catanzaro}]{shoeybi2019megatron}
Mohammad Shoeybi, Mostofa Patwary, Raul Puri, Patrick LeGresley, Jared Casper,
  and Bryan Catanzaro. 2019.
\newblock Megatron-lm: Training multi-billion parameter language models using
  gpu model parallelism.
\newblock \emph{arXiv preprint arXiv:1909.08053}.

\bibitem[{Socher et~al.(2013)Socher, Perelygin, Wu, Chuang, Manning, Ng, and
  Potts}]{socher2013recursive}
Richard Socher, Alex Perelygin, Jean Wu, Jason Chuang, Christopher~D Manning,
  Andrew~Y Ng, and Christopher Potts. 2013.
\newblock Recursive deep models for semantic compositionality over a sentiment
  treebank.
\newblock In \emph{Proceedings of the 2013 conference on empirical methods in
  natural language processing}, pages 1631--1642.

\bibitem[{Sukhbaatar et~al.(2019)Sukhbaatar, Grave, Lample, Jegou, and
  Joulin}]{sukhbaatar2019augmenting}
Sainbayar Sukhbaatar, Edouard Grave, Guillaume Lample, Herve Jegou, and Armand
  Joulin. 2019.
\newblock Augmenting self-attention with persistent memory.
\newblock \emph{arXiv preprint arXiv:1907.01470}.

\bibitem[{Sukhbaatar et~al.(2015)Sukhbaatar, Weston, Fergus
  et~al.}]{sukhbaatar2015end}
Sainbayar Sukhbaatar, Jason Weston, Rob Fergus, et~al. 2015.
\newblock End-to-end memory networks.
\newblock In \emph{Advances in neural information processing systems}, pages
  2440--2448.

\bibitem[{Sung et~al.(2017)Sung, Kim, Jo, Yang, Kim, Lausen, Kim, Lee, Kwak, Ha
  et~al.}]{sung2017nsml}
Nako Sung, Minkyu Kim, Hyunwoo Jo, Youngil Yang, Jingwoong Kim, Leonard Lausen,
  Youngkwan Kim, Gayoung Lee, Donghyun Kwak, Jung-Woo Ha, et~al. 2017.
\newblock Nsml: A machine learning platform that enables you to focus on your
  models.
\newblock \emph{arXiv preprint arXiv:1712.05902}.

\bibitem[{Vaswani et~al.(2017)Vaswani, Shazeer, Parmar, Uszkoreit, Jones,
  Gomez, Kaiser, and Polosukhin}]{vaswani2017attention}
Ashish Vaswani, Noam Shazeer, Niki Parmar, Jakob Uszkoreit, Llion Jones,
  Aidan~N Gomez, {\L}ukasz Kaiser, and Illia Polosukhin. 2017.
\newblock Attention is all you need.
\newblock In \emph{Advances in neural information processing systems}, pages
  5998--6008.

\bibitem[{Verga et~al.(2020)Verga, Sun, Soares, and Cohen}]{verga2020facts}
Pat Verga, Haitian Sun, Livio~Baldini Soares, and William~W Cohen. 2020.
\newblock Facts as experts: Adaptable and interpretable neural memory over
  symbolic knowledge.
\newblock \emph{arXiv preprint arXiv:2007.00849}.

\bibitem[{Wang et~al.(2018)Wang, Singh, Michael, Hill, Levy, and
  Bowman}]{wang2018glue}
Alex Wang, Amanpreet Singh, Julian Michael, Felix Hill, Omer Levy, and Samuel~R
  Bowman. 2018.
\newblock Glue: A multi-task benchmark and analysis platform for natural
  language understanding.
\newblock \emph{arXiv preprint arXiv:1804.07461}.

\bibitem[{Warstadt et~al.(2019)Warstadt, Singh, and
  Bowman}]{warstadt2019neural}
Alex Warstadt, Amanpreet Singh, and Samuel~R Bowman. 2019.
\newblock Neural network acceptability judgments.
\newblock \emph{Transactions of the Association for Computational Linguistics},
  7:625--641.

\bibitem[{Weston et~al.(2014)Weston, Chopra, and Bordes}]{weston2014memory}
Jason Weston, Sumit Chopra, and Antoine Bordes. 2014.
\newblock Memory networks.
\newblock \emph{arXiv preprint arXiv:1410.3916}.

\bibitem[{Williams et~al.(2017)Williams, Nangia, and
  Bowman}]{williams2017broad}
Adina Williams, Nikita Nangia, and Samuel~R Bowman. 2017.
\newblock A broad-coverage challenge corpus for sentence understanding through
  inference.
\newblock \emph{arXiv preprint arXiv:1704.05426}.

\bibitem[{Wolf et~al.(2019)Wolf, Debut, Sanh, Chaumond, Delangue, Moi, Cistac,
  Rault, Louf, Funtowicz et~al.}]{wolf2019transformers}
Thomas Wolf, Lysandre Debut, Victor Sanh, Julien Chaumond, Clement Delangue,
  Anthony Moi, Pierric Cistac, Tim Rault, R{\'e}mi Louf, Morgan Funtowicz,
  et~al. 2019.
\newblock Transformers: State-of-the-art natural language processing.
\newblock \emph{arXiv preprint arXiv:1910.03771}.

\bibitem[{Zhou et~al.(2020)Zhou, Ge, Xu, Wei, and Zhou}]{zhou2020scheduled}
Wangchunshu Zhou, Tao Ge, Ke~Xu, Furu Wei, and Ming Zhou. 2020.
\newblock Scheduled drophead: A regularization method for transformer models.
\newblock \emph{arXiv preprint arXiv:2004.13342}.

\bibitem[{Zhu et~al.(2015)Zhu, Kiros, Zemel, Salakhutdinov, Urtasun, Torralba,
  and Fidler}]{zhu2015aligning}
Yukun Zhu, Ryan Kiros, Rich Zemel, Ruslan Salakhutdinov, Raquel Urtasun,
  Antonio Torralba, and Sanja Fidler. 2015.
\newblock Aligning books and movies: Towards story-like visual explanations by
  watching movies and reading books.
\newblock In \emph{Proceedings of the IEEE international conference on computer
  vision}, pages 19--27.

\end{thebibliography}
